%% file: colm2026_paper.tex
\documentclass{article} 
\usepackage[preprint]{colm2026_conference}

\usepackage{microtype}
\usepackage{graphicx}
\usepackage{booktabs} 

\usepackage{hyperref}
\usepackage{url}

\usepackage{lineno}

\definecolor{darkblue}{rgb}{0, 0, 0.5}
\hypersetup{colorlinks=true, citecolor=darkblue, linkcolor=darkblue, urlcolor=darkblue}

\usepackage{xcolor}
\usepackage{color, colortbl}
\definecolor{Gray}{gray}{0.9}
\usepackage[normalem]{ulem} 

\newcommand{\myarxivtitle}{Learning to Rewrite Tool Descriptions for Reliable LLM-Agent Tool Use}

\usepackage{xspace}

\newcommand{\tracefree}{Trace-Free\xspace}
\newcommand{\tracefreemix}{Trace-Free+\xspace}

\usepackage[skins,breakable,listings]{tcolorbox}
\usepackage{listings}
\definecolor{lightgrey}{rgb}{0.827, 0.827, 0.827}

\usepackage[T1]{fontenc}
\usepackage[utf8]{inputenc}   
\usepackage{textcomp}

\usepackage{subcaption}

\usepackage{amsmath}
\usepackage{amssymb}
\usepackage{mathtools}
\usepackage{amsthm}

\usepackage[capitalize,noabbrev]{cleveref}

\theoremstyle{plain}

\theoremstyle{definition}

\theoremstyle{remark}

\usepackage[textsize=tiny]{todonotes}

\usepackage{enumitem}

\usepackage{multirow}

\usepackage{caption}
\title{\myarxivtitle}

\author{Ruocheng Guo\thanks{Equal contribution.} \quad Kaiwen Dong\footnotemark[1] \quad Xiang Gao \quad Kamalika Das \\
Intuit AI Research, Mountain View, CA, USA \\
\texttt{Kamalika\_Das@intuit.com}
}

\begin{document}

\ifcolmsubmission
\linenumbers
\fi

\maketitle

\input{abstract_colm}

\input{body/intro_colm}\vspace{-6pt}
\input{body/problem_statement_colm}\vspace{-6pt}
\input{body/method_colm}\vspace{-6pt}
\input{body/experiments_colm}
\vspace{-6pt}
\input{body/related_work}\vspace{-6pt}
\input{body/limitations}\vspace{-6pt}
\input{body/conclusion}


\bibliography{example_paper}
\bibliographystyle{colm2026_conference}

\appendix
\input{body/appendix_colm}

\end{document}

%% file: abstract_colm.tex
\begin{abstract}

While most efforts to improve LLM-based tool-using agents focus on the agent itself — through larger models, better prompting, or fine-tuning — agent performance increasingly plateaus due to the quality of the tool interfaces these agents consume.
Tool descriptions are often written for human developers and tolerate ambiguity that agents cannot resolve, particularly as the number of candidate tools grows.
Existing approaches to improving tool interfaces (1) require re-running a multi-stage per-tool pipeline --- synthesizing queries, executing an agent to collect trajectories, annotating trajectories, and prompting a strong LLM multiple times --- for every API that enters the catalog, and (2) typically optimize each tool independently, limiting scalability and generalization to unseen tools.
We propose \textbf{\tracefreemix}, a curriculum learning framework that progressively transfers supervision from trace-rich settings to trace-free deployment, encouraging the model to internalize reusable patterns of what makes a tool description effective.
To support this approach, we construct a large-scale dataset of high-quality tool interfaces derived from real-world APIs through a principled data synthesis workflow.
Experiments on widely adopted benchmarks show that Trace-Free+ improves robustness as tool catalogs scale to 150+ candidates — in scaling experiments, reducing accuracy degradation by 29.23\% and improving average query-level success by 60.89\% on StableToolBench — generalizes across domains without retraining, and provides complementary gains on top of agent fine-tuning.
\end{abstract}

%% file: body/intro_colm.tex
\section{Introduction}

Most efforts to improve LLM-based tool-using agents focus on the agent itself — stronger foundation models~\citep{team2025kimi,openai2024gpt4technicalreport,comanici2025gemini25pushingfrontier,yang2025qwen3}, better prompting~\citep{spiess2025autopdl,wu2024avatar}, or fine-tuning~\citep{dong2025agentic,qiwebrl}. As gains from scaling agents plateau, a fundamental bottleneck lies in the tool interfaces these agents consume.
Existing tool descriptions are written for human developers: they tolerate ambiguity, leave constraints implicit, and assume background knowledge that agents cannot acquire~\citep{hsieh2023tool}. As the number of candidate tools grows into the hundreds, these interface deficiencies compound — agents face not just harder reasoning but noisier decision surfaces where poorly specified tools become indistinguishable from relevant ones~\citep{qu_towards_2024}.


Consider a concrete example (see Fig.~\ref{fig:example_D2}): a scholarly API named \textit{publication\_year.find} with original description ``Fetches the year a particular scientific work was published.'' When asked about Newton's law of universal gravitation, an agent passes the colloquial phrase ``Law of Universal Gravitation'' as \texttt{work\_title} — a plausible but incorrect input. A description specifying that the API requires the ``full title'' leads the agent to the formal work title \textit{Philosophi\ae{} Naturalis Principia Mathematica}, producing a correct call.

{Approaches such as DRAFT~\citep{quexploration}, Play2Prompt~\citep{fang2025play2prompt} and D2 (Section~\ref{subsec:data-synthesis}) require running a full per-tool pipeline for every new API: synthesizing realistic queries, executing an agent to collect success/failure traces, annotating trajectories against ground truth, and prompting a strong LLM multiple times to refine the description. In enterprise settings where API catalogs change frequently, this pipeline must be re-executed for each new tool --- a recurring operational burden that is further compounded when traces cannot be collected due to cold-start, safety, or privacy constraints.} 

More critically, these methods optimize each tool in isolation: they do not learn transferable patterns of what makes a tool description effective, leading to poor generalization to unseen tools and degraded performance as candidate sets grow. Prompting-based methods such as EasyTool~\citep{yuan2025easytool} avoid trace dependence but similarly treat each tool independently — they cannot learn or transfer effective interface patterns across tools, and must re-invoke a strong LLM for every new tool at inference time.

%
%
A key hypothesis underlying our approach is that effective tool descriptions follow a bounded and reusable set of interface patterns (following the principle of information hiding~\citep{parnas1972criteria}). While the space of strategies an agent may learn to cope with arbitrary tools is effectively unbounded, the ways to specify a good interface—including scope definition, parameter constraints, output semantics, and dependency structure—are comparatively limited and recur across APIs.
This asymmetry suggests that interface optimization can be learned as a transferable capability: instead of adapting agents to each tool, we can learn to rewrite tools into a form that is consistently interpretable by agents.
%
We empirically validate this bounded-pattern hypothesis in Section~\ref{sec:trace-free-eval}, where we show that Trace-Free+ achieves up to 97.2\% pattern coverage across five categories on 4,585 unseen tools, while the original descriptions ($D_0$) cover fewer than 12\% in any category (Table~\ref{tab:pattern_coverage}).
Rather than replacing agent fine-tuning, better interfaces reduce how often fine-tuning is needed and amplify its effectiveness when applied.

We propose Trace-Free+, a framework that operationalizes this insight by treating interface optimization as a learned, transferable capability. Execution traces provide rich supervision for learning what makes a description effective, but are unavailable for new tools at deployment time. Trace-Free+ resolves this tension through curriculum learning~\citep{bengio2009curriculum}: training begins with trace-based examples (easier, more supervision) and gradually transitions to trace-free examples where descriptions must be generated from the tool schema alone (harder, matching deployment conditions). This enables the model to internalize patterns from trace-rich settings and apply them to unseen tools without requiring any tool interaction. Unlike the per-tool pipelines described above, onboarding a new tool requires only its schema as input, with no query synthesis, trace collection, annotation, or rule extraction. We instantiate this approach using a large-scale dataset of high-quality tool interfaces derived from real-world APIs via a principled synthesis pipeline.

Extensive experiments demonstrate that Trace-Free+ achieves state-of-the-art results on StableToolBench~\citep{guo2024stabletoolbench} and RestBench~\citep{song2023restgpt} in the trace-free setting, with all test tools unseen during training. Notably, in scaling experiments with up to 150+ candidate tools, it reduces accuracy degradation by 29.23\% and improves query-level success by 60.89\% on average, provides complementary gains on top of agent fine-tuning, and transfers to the Berkeley Function Calling Leaderboard (BFCLv2)~\citep{patil2025bfcl} — lifting the state-of-the-art Gemini-3-pro-preview~\citep{gemmateam2025gemma3technicalreport} by up to 1.4 points purely through better tool interfaces.

Our contributions are: (1) \tracefreemix, a curriculum learning framework transferring supervision from trace-rich training to trace-free deployment; (2) a large-scale dataset of high-quality tool interfaces derived from real-world APIs; and (3) state-of-the-art results on StableToolBench, RestBench, and BFCLv2 with all test tools unseen during training.


\begin{figure}[htbp]
    \centering
    \includegraphics[width=0.9\linewidth]{figs/example_newton_smaller.png}
    \caption{An illustration of the proposed tool interface improvement pipeline. Compared to the original description ($D_0$), the learned description generator produces more effective tool descriptions that lead to better tool usage.}
    \label{fig:example_D2}
\end{figure}

%% file: body/problem_statement_colm.tex
\section{Problem Statement}

We cast tool interface improvement as a supervised learning problem over a distribution of tools, training a description generator that internalizes effective interface patterns and transfers them to unseen tools at deployment time.

\textbf{Notation.}
A multi-step query decomposes into subtasks $h_t = (x_t, a_t, p_t, o_t)$, denoting the input $x_t$, tool $a_t$, parameters $p_t$, and output $o_t$ fed into $x_{t+1}$.
Each tool $a_i = \{d_i, s_i\} \in \mathcal{A}$ has description $d_i$ and schema $s_i$; only descriptions are improved (schemas are held fixed after preprocessing). Interface quality is measured by $R(\mathcal{A}; \mathcal{Q})$, instantiated as subtask- and query-level success rates (Section~\ref{subset:exp_setup}). From $\mathcal{A}_{tr}$ with queries $\mathcal{Q}_{tr}$, we aim to produce improved descriptions $d'_i$, optimizing $R(\mathcal{A}_{ts}; \mathcal{Q}_{ts})$ on held-out test tools $\mathcal{A}_{ts}$.

We evaluate three settings. \textbf{Trace-free}: given unseen tools $\mathcal{A}_{ts}$ without any execution, the generator produces improved interfaces from original tool — evaluated both \emph{in-domain} (StableToolBench) and \emph{cross-domain} (RestBench, BFCLv2, unseen API types). \textbf{Scaling}: candidate sets are augmented to 150+ tools, testing whether improved interfaces maintain their advantage as selection becomes harder. \textbf{Amplifying agent fine-tuning}: testing whether generated descriptions can provide additional gains on top of agent fine-tuning.

%% file: body/method_colm.tex
\section{Methodology}
\vspace{-4pt}

Our goal is to improve unseen tools at deployment time without re-running the full per-tool pipeline that high-quality description generation requires: synthesizing queries for new APIs, executing an agent to collect traces, annotating them, and prompting strong LLMs multiple times to refine each tool description. This pipeline must be repeated for every new tool --- a recurring operational cost that is prohibitive at scale and especially in cold-start, safety-critical, or privacy-constrained settings. To achieve deployment-time tool improvement, we train a model that internalizes what makes tool descriptions effective across a large set of tools and applies these patterns zero-shot from the schema alone. Existing methods cannot achieve this: they treat each tool independently and cannot learn cross-tool patterns, and trace-based refinement requires re-running the full pipeline for every new tool.
We present the learning framework that enables this transfer (Section~\ref{subsec:trace-free-plus}) and describe the data synthesis pipeline that produces the supervision for it (Section~\ref{subsec:data-synthesis}).
\vspace{-4pt}
\subsection{Trace-Free+: Curriculum Learning for Transferable Interface Optimization}
\label{subsec:trace-free-plus}
\vspace{-4pt}
A fundamental tension underlies tool interface improvement: execution traces provide the richest supervision for what makes a description effective---revealing valid tool use cases and parameter constraints---yet traces can be unavailable for new tools at deployment time due to cold-start, privacy, or safety constraints. The core question is therefore: \emph{can a model trained with trace-based data learn to generate effective descriptions without traces at inference?}

\textbf{Learning formulation and curriculum design.}
We cast tool interface improvement as a supervised learning problem over a distribution of tools. Given training tools $\mathcal{A}_{tr}$ and their improved descriptions $d'_i$ (Section~\ref{subsec:data-synthesis}), we fine-tune an open-weight LLM to serve as a description generator. Each training example pairs an input---consisting of the original tool interface $a_i$ and, optionally, a summary of execution traces $h_i = \text{Summary}(H(a_i))$---with the target improved description $d'_i$.

A na\"{i}ve approach would train exclusively on one type of samples. Training only with traces creates a mismatch: the model learns to condition on information absent at deployment. Training only without traces discards the richest supervision signal observable only through execution, such as which parameter formats cause errors or which tool combinations lead to ordering failure. Neither extreme allows the model to \emph{understand} how traces inform effective descriptions and \emph{abstract} those patterns for trace-free application.

Curriculum learning resolves this tension by structuring training to progressively bridge the gap between trace-rich supervision and trace-free deployment. Training begins with a higher proportion of trace-based examples, where the model learns the mapping from traces to effective descriptions---for instance, that a parameter named \texttt{ip\_address} should specify ``only IPv4 and IPv6 formats accepted.'' As training progresses, the proportion of trace-free examples increases until they dominate. In this phase, the model must generate the same quality of descriptions without traces, forcing it to internalize the \emph{types} of patterns that matter---tool selection scope, cross-tool dependencies, parameter constraints---rather than relying on explicit trace evidence, as illustrated in Fig.~\ref{fig:example_D2} and Table~\ref{tab:case-study}. We denote this curriculum-trained model as \textbf{Trace-Free+}. For controlled comparison, we also train \textbf{Trace-Free} (trace-free examples only, no curriculum).



The curriculum encourages the model to abstract reusable patterns from trace-rich examples and apply them in trace-free contexts. These patterns fall into five categories (Table~\ref{tab:pattern_category_def}, Appendix~\ref{appendix:case-study}): (1)~\emph{tool selection scope} --- when to use versus not use an API, and how it differs from similar tools; (2)~\emph{cross-tool dependencies} --- parameter values that must come from a specific upstream endpoint; (3)~\emph{output description} --- what fields and types the response contains or omits; (4)~\emph{parameter constraints} --- valid formats, ranges, and enumerated values; and (5)~\emph{cross-parameter dependencies} --- parameters that must be paired together or are mutually exclusive.
%
Because training spans hundreds of diverse tools, the model encounters recurring instances of all five patterns and learns to anticipate them on entirely unseen tools without any tool interaction. For example, $D_0$ of a Walk Score API states only "Get Walk Score" and copies its format field verbatim from an unrelated movie API ("Type of result to return: (movie, series, episode)").
From the schema alone, \tracefreemix infers that lat must be a decimal in [-90, 90], lon in [-180, 180], that bike and transit accept only the exact string '1' (not 'true' or 'on'), and that format must be 'json' or empty. It also adds scope exclusions: "Do not use for real-time traffic data or historical trends." None of these constraints were available in any execution trace — they were inferred purely from schema-level patterns learned during training. Additional examples appear in Table~\ref{tab:case-study} (Appendix~\ref{appendix:case-study}).

\begin{figure}[h]
    \centering
    \includegraphics[width=0.75\linewidth]{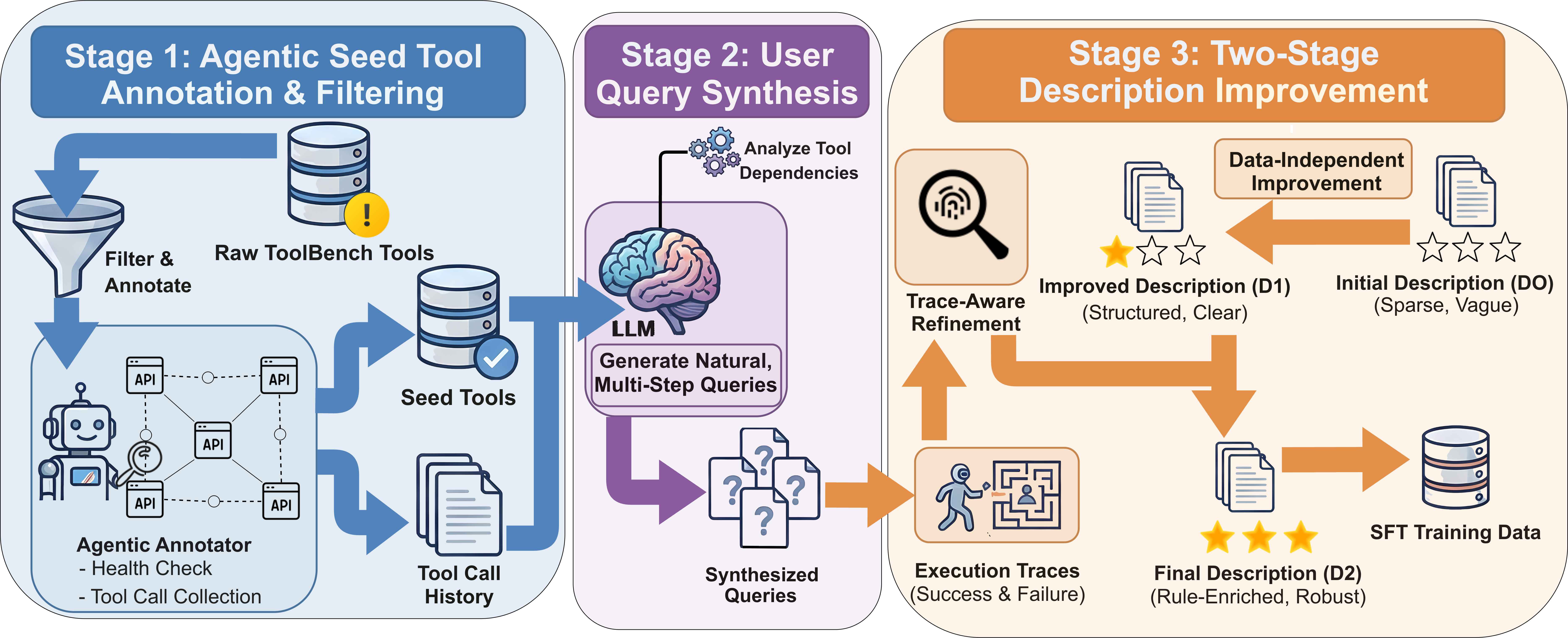}
    \caption{The data synthesis pipeline.}
    \label{fig:data_synthesis}\vspace{-4pt}
\end{figure}\vspace{-6pt}
\subsection{Data Synthesis for Tool Interface Improvement}

\label{subsec:data-synthesis}

We construct training data through a three-stage pipeline that converts real-world APIs into high-quality supervision for tool description generation. At a high level, we (1) collect working tool interfaces, (2) synthesize multi-step queries that expose interface deficiencies, and (3) generate improved descriptions that encode both general principles and trace-derived constraints.
To support this curriculum, we require training data that: (1) coverage across a diverse tool distribution, so the model encounters generalizable tool description patterns rather than memorizing tool-specific fixes, and (2) high-quality target descriptions that encode these generalizable patterns. The two types of samples share the same target but differ in input: trace-based samples provide the original description, parameter schema together with a trace summary, while trace-free examples do not provide any trace summary, forcing the model to internalize patterns that compensate for the absence of traces. We construct such data through a three-stage pipeline (Figure~\ref{fig:data_synthesis}) over real-world APIs. Full details, prompts, and examples are in Appendix~\ref{appendix:implementation}.

\textbf{Stage 1: Seed tool annotation and filtering.}
We source tools from ToolBench~\citep{qin_toolllm_2023}, spanning 49 categories of real-world RESTful APIs. An agentic annotator programmatically interacts with each provider to label endpoint health and record request--response examples, yielding 5,576 tools split into $\mathcal{A}_{ts}$ (4,585 tools appearing as candidates in StableToolBench test queries, held out entirely) and $\mathcal{A}_{tr}$ (991 remaining tools, 2,189 synthesized queries), ensuring no test tool is seen during training. Each $\mathcal{A}_{tr}$ tool contributes multiple trace-based training examples (one per synthesized query) plus one trace-free example, so the total number of training instances exceeds the number of tools; see Table~\ref{tab:synthetic_stats} and Appendix~\ref{appendix:implementation} for full statistics and filtering criteria.

\textbf{Stage 2: Dependency-aware query synthesis.}
Many interface deficiencies surface only through multi-step execution. We leverage API call histories from Stage~1 to identify inter-call dependencies: an LLM selects APIs forming a coherent workflow and generates a natural-language query requiring all selected APIs in sequence (details in Appendix~\ref{appendix:query_synthesis}).

\textbf{Stage 3: Two-stage description improvement.}
We run a tool-using agent on the synthesized queries and save successful and failed trajectories. Starting from original descriptions $D_0$, we first apply general documentation guidelines---specifying use cases and parameter constraints---to produce $D_1$. We then refine $D_1$ with general rules extracted from failure traces via RIMRULE~\citep{gao2025rimrule}---e.g., acceptable value formats and undocumented preconditions---yielding $D_2$. The resulting $D_2$ descriptions serve as the supervision target for both curriculum branches; what differs is the input the model receives at training time.

While $D_2$ is generated per-tool, \tracefreemix trained on these examples across hundreds of tools abstracts cross-tool patterns---when to use, recurring parameter constraints, output format, effective documentation strategies---and applies them zero-shot to unseen tools. At the same time, \tracefreemix does not aim to outperform $D_2$, but to match its quality without incurring its per-tool cost, enabling scalable deployment in settings where running $D_2$ is infeasible. Per-tool prompting methods~\citep{yuan2025easytool,quexploration,fang2025play2prompt} cannot leverage such patterns because they improve each tool in isolation.

%% file: body/experiments_colm.tex
\section{Experiments}
\vspace{-2pt}
\label{sec:exp}
We evaluate \tracefreemix across the following dimensions:
(1)~\textbf{trace-free generalization}---whether a model trained on one set of tools can produce effective descriptions for unseen tools without traces, including \textbf{cross-domain transfer} to tools and benchmarks outside the training distribution (see Section~\ref{sec:trace-free-eval});
(2)~\textbf{scaling robustness}---whether improved descriptions maintain their advantage as candidate sets grow to 150+ tools;
(3)~\textbf{amplifying agent fine-tuning}---whether our method provides additional gains on top of agent fine-tuning.
In all settings, \tracefreemix and its variants are evaluated on \emph{tools unseen during training}. 
%
\vspace{-4pt}
\subsection{Experimental Setup}
\label{subset:exp_setup}
\vspace{-4pt}
\paragraph{Benchmarks.}
Our in-domain benchmark is StableToolBench~\citep{guo2024stabletoolbench} with six subsets. They have single-step (G1) and multi-step (G2--G3) queries of increasing difficulty, totaling 764 solvable queries over 4,585 candidate tools ($\mathcal{A}_{ts}$). The training set $\mathcal{A}_{tr}$ comprises 991 tools with 2,189 synthesized queries.
Following~\citep{lu2025codetool}, we correct a subset of parameter schemas that are inconsistent with server requirements (Appendix~\ref{appendix:stb_preprocess}).
For cross-domain transfer, we test on RestBench~\citep{song2023restgpt} --- TMDB (100 queries, 54 tools) and Spotify (57 queries, 40 tools) --- and BFCLv2~\citep{patil2025bfcl} (1,390 Non-Live and 2,251 Live instances), both out-of-domain. Full benchmark statistics are in Table~\ref{tab:statistics}.

\textbf{Baselines and agents.}
We focus on the \textbf{trace-free} setting, where we compare against the original descriptions ($D_0$), the prompting-improved descriptions ($D_1$), and EasyTool~\citep{yuan2025easytool}.
Our primary tool-using agent is GPT-4.1; to test whether improved descriptions generalize across agents, we additionally evaluate with Qwen3-4B-Instruct~\citep{yang2025qwen3} in both its base and fine-tuned variants.

\textbf{Evaluation protocol.}
We introduce step-wise teacher-forcing evaluation to isolate the effect of tool descriptions from compounding execution errors.
At each step, the ground-truth API is called regardless of the agent's selection, ensuring that subsequent steps receive correct intermediate context.
This design guarantees that any subtask failure reflects the agent's misunderstanding of the current tool's description rather than corrupted context from earlier mistakes, enabling cleaner error attribution across methods. In addition, to better reflect real-world deployment, the scaling experiments use a non-teacher-forcing setting where the agent's own selected tool is executed at each step.
We report \emph{subtask-level (SL)} and \emph{query-level (QL)} success rates: a subtask succeeds if the correct tool is selected and execution completes successfully; a query succeeds iff all its subtasks succeed. Both metrics are based on ground truth and do not rely on LLM-as-a-judge. 
\vspace{-4pt}

\subsection{Trace-free Evaluation}
\label{sec:trace-free-eval}

The trace-free evaluation tests generalizability: the trained generator is applied to unseen $\mathcal{A}_{\text{ts}}$ without traces, with GPT-4.1 as the primary agent. Cross-domain evaluation covers RestBench~\citep{song2023restgpt} (TMDB and Spotify) and BFCLv2~\citep{patil2025bfcl}, with Claude Sonnet 4.5 and Gemini-3-pro-preview additionally included for BFCLv2 to test agent-agnostic transfer. Results with Qwen3-4B-Instruct and agent fine-tuning are in Section~\ref{subsec:agent-ft}. 

\textbf{In-domain results.}
Table~\ref{tab:stb-trace-free-split} reveals several patterns. First, Trace-Free+ significantly
outperforms Trace-Free across most subsets, confirming that curriculum learning---which
exposes the model to trace-based supervision before
transitioning to trace-free generation---transfers
knowledge that pure trace-free training cannot
acquire.


Second, the split averages reveal where description
quality matters most. On multi-step queries
(G2+G3), \tracefreemix achieves 44.6 QL, improving
over $D_0$ by 11.1 points and over $D_1$ by 3.1
points. On single-step queries (G1), \tracefreemix
is within 0.9 QL of $D_1$ (60.7 vs.\ 61.6). The
G1 gap partly reflects a training data distribution
choice: our synthesized queries require 3 tools on
average, prioritizing practical multi-step agentic
settings at the cost of underrepresenting
single-step patterns. We leave augmenting training
data with single-step queries as future work.
In practice, \tracefreemix is the choice for
large catalogs or multi-step workflows while $D_1$ can be used for single-step queries.

Third, the multi-step advantage stems from
\tracefreemix's ability to internalize cross-tool
patterns that $D_1$ cannot capture. $D_1$ applies
fixed heuristics independently per tool, relying
on whatever the original schema says about
inter-API relationships. \tracefreemix, by
contrast, has learned from real execution traces
across hundreds of training tools which APIs
produce outputs that feed into others, what
parameter formats cause failures, and how scope
boundaries interact across endpoints. Multi-hop
queries amplify this difference: an imprecise
constraint at step~1 cascades to all subsequent
steps, compounding $D_1$'s per-tool blind spots.
Our case study
(Table~\ref{tab:pattern_coverage},
Appendix~\ref{appendix:case-study}) confirms this
mechanistically: \tracefreemix achieves higher
parameter constraint coverage (94.2\% vs.\ 87.9\%)
and cross-parameter dependency coverage
(17.1\% vs.\ 8.5\%) than $D_1$---the two
categories most directly responsible for correct
argument construction in multi-step chains.
%
EasyTool performs below $D_0$ because its manually optimized prompts were designed for older models (ChatGPT, Vicuna-30B) and do not transfer to the newer agents used here, consistent with results in~\citet{fang2025play2prompt}.
To empirically validate the concentrated set of reusable interface patterns, we classify all descriptions on $\mathcal{A}_{ts}$ tools (Table~\ref{tab:pattern_coverage}, Appendix~\ref{appendix:case-study}). Original descriptions ($D_0$) cover fewer than 12\% of tools in any of the five pattern categories. Trace-Free+ raises coverage to 97.2\% for tool selection scope and 94.2\% for parameter constraints — the two categories directly responsible for correct tool selection and execution — while also reaching 30.0\% for cross-tool dependencies and 17.1\% for cross-parameter dependencies (vs. $D_1$'s 8.5\%).
$D_1$ achieves near-complete output description coverage (98.6\%) because its template guidelines explicitly enumerate output fields, whereas Trace-Free+ encodes output information implicitly; this gap does not substantially affect downstream performance because output description primarily helps agents interpret responses rather than select tools or construct arguments. While the static performance gap between Trace-Free+ and $D_1$ on single-step queries is modest, \tracefreemix outperforms $D_1$ substantially as catalogs scale, with the widening concentrated on multi-step queries where description quality compounds across steps (Section~\ref{subsec:scaling}). A parameter study on the trace-free data ratio is reported in Appendix~\ref{appendix:add_exp_results}.

\textbf{Cross-domain results on RestBench and BFCLv2.} We test whether the learned patterns transfer
beyond the training domain. Models fine-tuned on $\mathcal{A}_{\text{tr}}$ of
StableToolBench are
evaluated on the TMDB and Spotify datasets of
RestBench~\citep{song2023restgpt}, with all API
tools from each dataset included as candidates.
As shown in Table~\ref{tab:restbench}, Trace-Free+ outperforms all baselines by a substantial margin, achieving up to
+51.3\% relative improvement over $D_0$ on
TMDB query-level success rate and +41.3\% on Spotify.
These gains reflect the domain-agnostic nature of effective interface patterns: with APIs from an entirely unseen benchmark, the underlying improvements---clearer scope boundaries, explicit parameter constraints, disambiguation of overlapping endpoints---transfer directly without any retraining.


To further stress-test cross-domain transfer, we
apply Trace-Free+ to the BFCLv2~\citep{patil2025bfcl} as the number of candidate tools per query is large enough. It is a benchmark that evaluates function calls using AST-based verification---a fundamentally different evaluation paradigm from the teacher-forcing protocol used above. We evaluate on both Non-Live and Live splits using
three strong proprietary models: GPT-4.1, Claude Sonnet 4.5, and Gemini-3-pro-preview. As shown in
Table~\ref{tab:bfcl-transfer} (Appendix~\ref{appendix:add_exp_results}), Trace-Free+
consistently improves all three models on both
splits purely through better tool interfaces,
without any modification to the agent models
themselves. The largest absolute gain is
on Gemini-3-pro-preview Live (1.68\% relative improvement, or +1.43 points),
lifting the state-of-the-art to 86.41\%. Claude Sonnet 4.5 benefits the most in relative terms (3.79\% relative improvement, or +2.39 absolute points)). These results confirm that tool interface optimization is model-agnostic: the same descriptions improve GPT-4.1, Claude Sonnet 4.5, and Gemini-3-pro-preview without modification to any agent.

%


\input{tables/trace_free_stb_0310_split_avg}

\input{tables/trace_free_restbench}

\subsection{Scaling Experiments}
\label{subsec:scaling}
In practice, agents are routinely exposed to large, uncurated tool catalogs---yet performance degrades sharply as the candidate pool grows, since poorly specified descriptions become indistinguishable and selection errors compound across steps~\citep{qu_towards_2024}. Most benchmarks, including StableToolBench, evaluate on small curated tool sets that mask this failure mode; here, we directly test robustness as candidate sets scale to 150+ tools.

For this setting, we augment each query in StableToolBench with additional tool candidates.
Specifically, we consider three types of additional tool candidates: (1) relevant tools from the same category, (2) relevant tools from other categories, (3) random APIs from other categories. We let 10\% of the candidate set be type (1) and (2) and 90\% be type (3) to avoid making API selection too hard, resulting in small difference in performance across descriptions.
%
%
Unlike prior scaling studies~\citep{qin_toolllm_2023,yuan2025easytool,quexploration} that evaluate only the effectiveness of tool retrievers as a separate stage, we directly expose the full candidate set to agents and measure the performance end-to-end. This setting better reflects practical usage and avoids a fixed retrieval stage prior to agent execution, which is increasingly unnecessary given the large context windows supported by modern LLMs. For example GPT-4.1 and Gemini 2.5 Flash support context windows of up to one million tokens.
To better reflect realistic deployment, we use a non-teacher-forcing setting for scaling experiments where the agent's selected tool is executed at each step and descriptions are penalized if the number of decomposed tool calls differs from the ground truth.

\begin{figure*}[htbp]
  \centering

  \begin{minipage}[b]{0.99\linewidth}
    \centering
    \includegraphics[width=\linewidth]{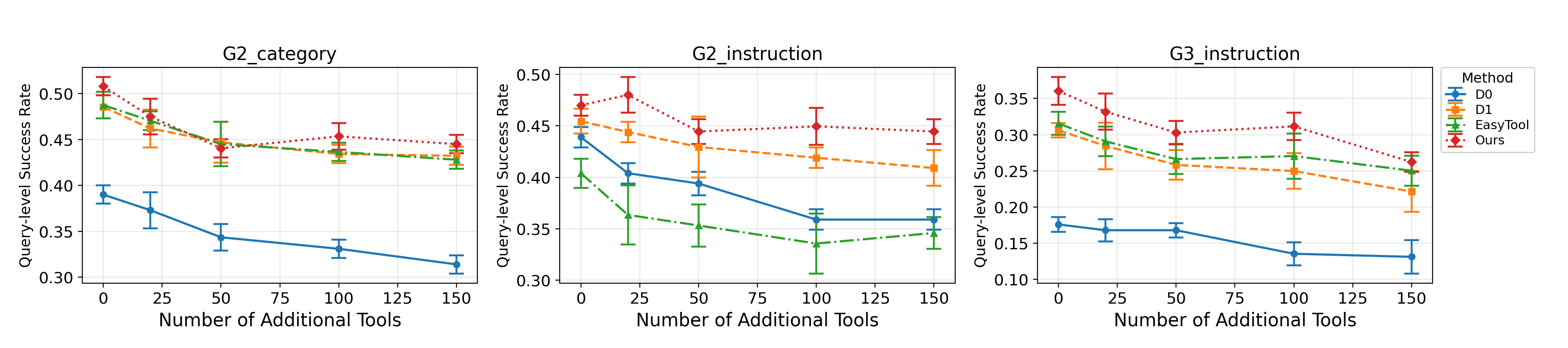}
  \end{minipage}

  \caption{Scaling experiment results on the more challenging G2-G3 subsets of StableToolBench. We report query-level (QL) results.}
  \label{fig:scaling_G3}
\end{figure*}

Fig.~\ref{fig:scaling_G3} shows that \tracefreemix outperforms the baselines and is more robust against an increasing number of additional APIs across the three multi-step subsets of StableToolBench.
From 0--150 additional tools, \tracefreemix reduces performance degradation of $D_0$ by 29.23\% and improves over $D_0$ by 60.89\% on average. This robustness stems from description quality: well-specified scope boundaries and parameter constraints help agents filter signal from noise as the candidate pool grows---a structural advantage that per-tool rewriting methods cannot provide. Notably, these are the largest gains observed in our evaluation, suggesting that interface optimization is most valuable when tool catalogs are large. This scalability advantage is unique to learned optimization: $D_1$ applies fixed heuristics per tool independently and cannot capture cross-tool patterns, whereas the SFT model internalizes recurring effective interface patterns from hundreds of training tools---enabling it to capture patterns such as cross-parameter dependencies and implicit constraints that are not explicitly encoded in $D_1$'s rule set---patterns that compound across steps as catalogs grow.
%

\vspace{-4pt}
\vspace{-8pt}

\vspace{-4pt}
\subsection{Amplifying Agent Fine-tuning}
\label{subsec:agent-ft}

\input{tables/finetuned_and_bfcl}

Interface optimization is not a replacement for agent fine-tuning---rather, by fixing the tool interface layer first, it reduces the burden on the agent and acts as a force multiplier when fine-tuning is applied. We fine-tune Qwen3-4B-Instruct on $\mathcal{A}_{tr}$ of StableToolBench and evaluate all combinations of base/fine-tuned agent with $D_0$/\tracefreemix descriptions on both StableToolBench and RestBench; Table~\ref{tab:finetuned-agent-summary} reports average query-level success rates.

On StableToolBench, agent fine-tuning and \tracefreemix yield individually comparable gains (\textbf{+10.7\% each over $D_0$}). Combining the two further improves performance to \textbf{41.8\%}, a \textbf{+11.7\%} relative gain over the baseline, indicating that description improvement and agent fine-tuning capture complementary aspects of tool-use competence.

The amplification is more pronounced in the cross-domain setting. On RestBench, agent fine-tuning brings the baseline from 54.1\% to only 55.0\%, reflecting limited cross-domain transfer---the agent adapts to StableToolBench's tool distribution but cannot generalize to RestBench's different API vocabulary. In contrast, \tracefreemix improves the base model to \textbf{62.1\% (+14.8\% relative)}, and combining both strategies reaches \textbf{62.9\% (+16.3\% relative)}, since interface patterns (scope boundaries, parameter constraints) are domain-agnostic by nature. Together, these results confirm that interface optimization and agent fine-tuning address complementary failure modes, with the combination most impactful in cross-domain deployment where fine-tuning alone is insufficient.

%% file: tables/trace_free_stb_0310_split_avg.tex
\begin{table*}[h]
\centering
\caption{Trace-free evaluation on StableToolBench for multi-step (G2+G3) and single-step (G1) subsets. SL/QL: subtask/query-level success rate. GPT-4.1 is the tool-using agent.}
\label{tab:stb-trace-free-split}
\setlength{\tabcolsep}{2.5pt}
\tiny
\resizebox{1\linewidth}{!}{
\begin{tabular}{lcccccccc@{\hspace{8pt}}cccccccc}
\toprule
& \multicolumn{8}{c}{\textbf{Multi-Step (G2+G3)}}
& \multicolumn{8}{c}{\textbf{Single-Step (G1)}} \\
\cmidrule(lr){2-9} \cmidrule(lr){10-17}

& \multicolumn{2}{c}{G2 Category}
& \multicolumn{2}{c}{G2 Instruction}
& \multicolumn{2}{c}{G3 Instruction}
& \multicolumn{2}{c@{\hspace{8pt}}}{\textbf{Avg}}
& \multicolumn{2}{c}{G1 Category}
& \multicolumn{2}{c}{G1 Instruction}
& \multicolumn{2}{c}{G1 Tool}
& \multicolumn{2}{c}{\textbf{Avg}} \\
\cmidrule(lr){2-3} \cmidrule(lr){4-5} \cmidrule(lr){6-7} \cmidrule(lr){8-9}
\cmidrule(lr){10-11} \cmidrule(lr){12-13} \cmidrule(lr){14-15} \cmidrule(lr){16-17}

& SL & QL
& SL & QL
& SL & QL
& SL & QL
& SL & QL
& SL & QL
& SL & QL
& SL & QL \\
\midrule

\tracefreemix
& \textbf{68.7 ± 0.7} & \textbf{50.8 ± 0.0}
& \textbf{71.4 ± 1.1} & \textbf{47.0 ± 0.1}
& \textbf{63.9 ± 1.8} & \underline{36.1 ± 0.1}
& \textbf{68.0} & \textbf{44.6}
& \underline{73.8 ± 0.7} & \textbf{65.6 ± 1.0}
& 72.6 ± 1.5 & 60.0 ± 1.0
& \underline{70.0 ± 1.4} & \textbf{56.4 ± 1.6}
& 72.1 & \underline{60.7} \\

\tracefree
& 66.4 ± 1.2 & 44.1 ± 0.0
& \underline{69.2 ± 4.0} & \underline{46.4 ± 3.6}
& \underline{59.9 ± 1.2} & \textbf{41.8 ± 1.2}
& \underline{65.2} & \underline{44.1}
& 71.8 ± 2.3 & 62.7 ± 2.1
& 70.8 ± 0.1 & 60.8 ± 0.9
& 68.5 ± 2.7 & 53.6 ± 2.8
& 70.4 & 59.0 \\

D1
& 67.9 ± 1.2 & \underline{48.5 ± 0.1}
& 67.4 ± 1.0 & 45.5 ± 0.1
& 45.2 ± 4.2 & 30.6 ± 0.1
& 60.2 & 41.5
& \textbf{75.5 ± 1.3} & \underline{64.9 ± 1.3}
& \textbf{74.7 ± 0.7} & \textbf{66.1 ± 0.6}
& 68.0 ± 0.6 & \underline{53.7 ± 0.5}
& \textbf{72.7} & \textbf{61.6} \\

D0
& \underline{68.4 ± 0.3} & 39.0 ± 0.0
& 67.8 ± 0.8 & 43.9 ± 0.1
& 50.7 ± 0.2 & 17.6 ± 0.1
& 62.3 & 33.5
& 73.0 ± 0.3 & 62.4 ± 1.5
& \underline{72.8 ± 0.4} & \underline{62.3 ± 0.9}
& \textbf{71.0 ± 0.9} & 52.8 ± 1.1
& \underline{72.3} & 59.2 \\

EasyTool
& 68.1 ± 0.7 & 40.4 ± 1.3
& 68.2 ± 0.6 & 40.4 ± 0.1
& 41.4 ± 1.7 & 31.6 ± 0.2
& 59.2 & 37.5
& 67.4 ± 1.1 & 56.9 ± 0.7
& 69.5 ± 1.2 & 56.0 ± 2.3
& 66.4 ± 1.3 & 48.7 ± 0.2
& 67.8 & 53.9 \\

\bottomrule
\end{tabular}\vspace{-12pt}
}
\end{table*}

%% file: tables/trace_free_restbench.tex
\begin{table}[t]
\centering
\caption{Trace-free evaluation results on RestBench -- TMDB and Spotify. SL: subtask-level, QL: query-level. No tool execution traces are available at inference
time.}
\label{tab:restbench}
\scriptsize
\setlength{\tabcolsep}{8pt}
\begin{tabular}{lcccc}
\toprule
& \multicolumn{4}{c}{\textbf{RestBench}} \\
\cmidrule(lr){2-5}

& \multicolumn{2}{c}{TMDB}
& \multicolumn{2}{c}{Spotify} \\
\cmidrule(lr){2-3} \cmidrule(lr){4-5}

\textbf{Method}
& SL & QL
& SL & QL \\
\midrule

\tracefreemix   
& \textbf{88.1 ± 0.4} 
& \textbf{74.9 ± 0.5} 
& \textbf{68.1 ± 0.3} 
& \textbf{49.3 ± 0.6} \\

\tracefree  
& \underline{78.4 ± 1.1} 
& 57.7 ± 1.2 
& 65.0 ± 1.6 
& 44.7 ± 2.8 \\

D1   
& 78.2 ± 0.1 
& \underline{58.0 ± 0.8} 
& \underline{65.1 ± 0.8} 
& \underline{45.7 ± 1.8} \\

D0              
& 69.8 ± 0.2 
& 49.5 ± 0.2 
& 57.1 ± 2.9 
& 34.9 ± 2.1 \\

EasyTool        
& 76.4 ± 0.1 
& 52.5 ± 0.0 
& 63.4 ± 0.8 
& 43.2 ± 0.2 \\

\bottomrule
\end{tabular}
\end{table}

%% file: tables/finetuned_and_bfcl.tex
\begin{table}[t]
    \centering
    \caption{Query-level success rate with base and fine-tuned Qwen3-4B-Instruct agents.}
    \label{tab:finetuned-agent-summary}
    \scriptsize
    \begin{tabular}{lcc}
    \toprule
    \textbf{Setting} & \textbf{StableToolBench} & \textbf{RestBench} \\
    \midrule
    Base + D0 & 37.4 & 54.1 \\
    Base + \tracefreemix & 40.1 (+ 10.7\%) & 62.1 (+ 14.8\%) \\
    Fine-tuned + D0 & 40.0 (+ 10.7\%) & 55.0 (+ 1.0\%) \\
    Fine-tuned + \tracefreemix & \textbf{41.8 (+ 11.7\%)} & \textbf{62.9 (+ 16.3\%)} \\
    \bottomrule
    \end{tabular}
    \end{table}

%% file: body/related_work.tex
\section{Related Work}
\vspace{-4pt}
\textbf{Tool-using LLM Agents.} Tool-using agents combine LLMs~\citep{openai2024gpt4technicalreport,comanici2025gemini25pushingfrontier,yang2025qwen3,team2025kimi} with external tools to extend capabilities beyond text generation~\citep{qin_toolllm_2023,schick2023toolformer,wang2025function}. The LLM serves as the controller~\citep{yao2023react}, deciding when and how to invoke tools. With external tools, an LLM can retrieve up-to-date information, perform calculations, and interact with external services~\citep{huang2025biomni,he2025pasa}. Early work such as Gorilla~\citep{patil2024gorilla} demonstrated that LLMs can be trained to call massive API sets, while ToolLLM~\citep{qin_toolllm_2023} extended this to 16,000+ real-world APIs. We focus on domain-specific APIs (e.g., from RapidAPI, TMDB, and Spotify), which require structured input arguments and return domain-specific outputs — making correct selection and execution more challenging than general-purpose tools, and making accurate tool descriptions especially critical.

\textbf{Tool Interface Improvement.}
Tool interfaces are important in guiding agents in tool selection and usage~\citep{xu2023tool,hsieh2023tool,bandlamudi2025framework,chen2025enhancing,faghih2025gaming,wolflein2025llm}.
A complementary line of work improves agents themselves via fine-tuning or contrastive reasoning~\citep{wu2024avatar,dong2025agentic}; our approach is orthogonal, targeting the interface layer rather than the agent.
Prompting-based methods have shown promising results: EasyTool~\citep{yuan2025easytool} addresses inconsistency, redundancy, and incompleteness via a two-step rewriting workflow; Play2Prompt~\citep{fang2025play2prompt} uses single-hop execution traces to improve descriptions with a strong LLM; and DRAFT~\citep{quexploration} iteratively collects traces and applies LLM self-correction to revise interfaces.
However, all three rely on per-tool trace collection and optimize each tool independently, preventing them from learning generalizable patterns across tools. DRAFT and Play2Prompt additionally cannot handle unseen tools without execution traces.

%% file: body/limitations.tex
\vspace{-4pt}
\section{Limitations}
\vspace{-4pt}
\textbf{Base model for description generation.} Our experiments use Qwen3-4B-Instruct as the description generator for its strong instruction-following ability. We leave extension to larger open-weight models as future work. Improvements to the base model are orthogonal to our contributions --- the curriculum learning framework and the large-scale dataset. A stronger base model may improve description quality further without replacing either.

\textbf{Single-step query performance.} \tracefreemix does not outperform $D_1$ on the single-step G1 subset of StableToolBench. This is a consequence of a training data distribution choice: our synthesized queries require 3 tools on average, prioritizing the practical agentic system at the cost of underrepresenting single-step patterns. As a result, $D_1$'s data-independent guidelines, which are applied uniformly regardless of query complexity, capture most of the available headroom on G1 while $D_0$ is already near the ceiling. Augmenting the training set with single-step queries is a direction for closing this gap (Section~\ref{subsec:scaling}).

\textbf{Precision of inferred constraints.} It is challenging to systematically measure the precision of constraints generated by \tracefreemix. While the case studies in Appendix~\ref{appendix:case-study} show correct inferences across diverse constraint types, the model may occasionally hallucinate constraints for unfamiliar APIs --- for instance, inventing value ranges not enforced by the server. Quantifying hallucination rates across constraint categories and developing verification mechanisms are important directions for future work.

%% file: body/conclusion.tex
\section{Conclusion}
\vspace{-2pt}
Our results suggest that effective tool descriptions follow learnable, transferable patterns that existing per-tool methods cannot exploit. We introduce \tracefreemix, a curriculum learning framework that transfers supervision from trace-rich training to trace-free deployment, paired with a large-scale dataset of high-quality interfaces derived from real-world APIs. Experiments show that in scaling experiments with up to 150+ candidate tools, \tracefreemix reduces accuracy degradation by 29.23\% and improves query-level success by 60.89\%, generalizes to unseen domains (RestBench, BFCLv2) without retraining, and acts as a force multiplier alongside agent fine-tuning — with the largest standalone gains appearing in cross-domain settings where fine-tuning alone shows limited transfer. These results suggest that interface quality is an under-exploited axis of agent improvement, and that learned interface optimization can complement advances in model capability and agent training. For practitioners, this yields a deployment strategy: $D_1$ suffices for simple queries, small-catalog settings, while \tracefreemix is better for large tool catalogs with multi-step workflows.

%% file: body/appendix_colm.tex
\appendix
\onecolumn
\section{Implementation Details}

\label{appendix:implementation}

\subsection{Detailed Agentic Tool Annotator}
The prompt of the agentic tool annotator can be found below.
\input{body/prompts/annotator}

\subsection{Detailed User Query Synthesis Procedure}
\label{appendix:query_synthesis}

This section provides a detailed description of the user query synthesis process.

\paragraph{Target Query Properties.}
High-quality synthetic queries must satisfy three properties. First, queries should sound natural and reflect how real users describe tasks, rather than exposing explicit tool usage or step-by-step instructions. Second, queries must require multiple tool calls to complete, such that no single API invocation suffices. Third, the required tool calls must exhibit dependency relationships, where later calls depend on the outputs of earlier ones, enforcing non-trivial planning and intermediate result handling.

\paragraph{Base Pipeline.}
We partially adopt the query synthesis pipeline from TOUCAN~\cite{xu2025toucan}, which emphasizes realism, linguistic quality, and multi-tool reasoning. Similar to TOUCAN, we prompt an LLM with tool schemas and descriptions to generate candidate queries under constraints that exclude trivial or single-step tasks. However, our approach differs in how tool combinations are selected and how dependencies are enforced.

\paragraph{Dependency-Aware Query Construction.}
In addition to schema information, we leverage API calling histories collected during seed tool annotation, which reveal common call orders, data flow patterns, and functional relationships between APIs within the same provider. We explicitly prompt the LLM to analyze these relationships before generating queries.

Concretely, for each API provider, the LLM is instructed to select three APIs whose functionalities exhibit clear dependency structure, such as retrieval followed by transformation or filtering followed by aggregation. The model first produces a brief dependency analysis describing how these APIs interact, and then generates a single user query that implicitly requires invoking all selected APIs in the correct order. Tool names and execution details are omitted from the query text to preserve naturalness.

\paragraph{Outcome.}
By grounding query synthesis in real API usage traces and explicit dependency reasoning, this process produces queries that are both linguistically natural and structurally challenging. These queries reliably induce multi-step tool-use trajectories with meaningful inter-call dependencies, which are critical for supervising and evaluating advanced tool-using LLM agents.

\subsection{Detailed Traces and Improved Description Generation}
\label{appendix:trace_desc_appendix}

This section provides a detailed description of how execution traces are collected and how they are used to generate improved tool descriptions $D1$ and $D2$.

\paragraph{Trace Collection.}
The synthesized user queries are designed to require multi-step tool use with explicit dependencies between APIs. We execute a tool-using agent on these queries and record full execution traces, including intermediate reasoning steps, tool calls, tool responses, and termination states. For each query, we retain both successful traces and failure traces, where failures include incorrect tool selection, invalid argument construction, premature termination, or unrecoverable tool errors.

We associate each failure trace with its corresponding ground-truth tool sequence, enabling direct comparison between incorrect and correct executions. This comparison allows us to identify whether failures arise from missing information in the tool description, unclear argument semantics, or undocumented usage constraints.

\paragraph{Data-Independent Description Improvement.}
Starting from the original tool description $D_0$, we first generate a data-independent improved description $D_1$. This step applies general guidelines for tool description writing, including: clearly stating the tool’s intent, specifying required versus optional parameters, documenting expected input formats, describing output semantics, and clarifying common error conditions. These guidelines are initialized from publicly available tool-use best practices\footnote{\url{https://platform.claude.com/docs/en/agents-and-tools/tool-use/implement-tool-use\#best-practices-for-tool-definitions}} and iteratively refined by measuring downstream agent performance when consuming $D_1$. The prompt used to generate $D_1$ is shown below.


\paragraph{Trace-Driven Rule Extraction.}
To incorporate execution-specific information, we further refine descriptions using rules extracted from traces. We adopt the RIMRULE framework~\cite{gao2025rimrule}, which compares failed traces against their corresponding ground-truth executions to identify root-cause reasoning errors. These errors are distilled into compact, generalizable rules that describe correct tool usage under specific conditions, such as required call ordering, necessary preconditions, or constraints on argument construction. The resulting rules form a reusable rule library derived from observed agent behavior.

\paragraph{Trace-Aware Description Generation.}
For each tool, we retrieve the subset of rules relevant to that tool and combine them with its $D_1$ description to generate a final description $D_2$. Unlike $D_1$, which is independent of execution context, $D_2$ explicitly encodes behavioral constraints grounded in observed failures and successes. This process produces descriptions that are tailored to the actual usage patterns of each tool while remaining general enough to apply across different queries.

The final $D_2$ descriptions are used as supervision for supervised fine-tuning of the description generator. 


\section{Experiment Setup Details}
\label{appendix:exp_details}

Here, we present additional details about the experiments for better understanding and reproducibility.

Table~\ref{tab:statistics} reports descriptive statistics for the three evaluation benchmarks used in our experiments. For StableToolBench, we report only solvable queries as defined by~\citet{guo2024stabletoolbench}.

\input{tables/dataset_description}

Table~\ref{tab:synthetic_stats} summarizes the synthesized SFT dataset described in Section~\ref{subsec:data-synthesis}. $\mathcal{A}_{ts}$ contains all tools appearing as candidates in StableToolBench test queries and is held out entirely during training; $\mathcal{A}_{tr}$ comprises the remaining tools used to produce training examples.

\input{tables/synthetic_data_stats}

\subsection{Teacher-forcing Evaluation}
\label{appendix:eval_protocol_metrics}
The teacher-forcing evaluation begins with a task decomposition step, adopted from DRAFT~\cite{qu_towards_2024}, to obtain a set of subtasks and their dependencies given a query. Each subtask requires no more than one tool to solve.
Then, for each subtask, we perform the following steps: subtask-level tool selection annotation, tool selection, tool execution, and tool response processing.
Subtask-level evaluation metrics are computed based on the results of tool selection and tool execution.
For the concern of budget, we use GPT-4.1 (2025-05-14) as our tool-using agent in all the experiments.

\paragraph{Subtask Tool Selection Annotation.} For calculation of tool selection accuracy on subtask and query-level and F1 score on tool-level, we annotated the ground truth API tool for each subtask using GPT-4.1, which also judges whether a subtask needs a tool or not. We manually checked correctness on 100 randomly selected subtasks; the annotation is correct in 96 cases.
The prompt of subtask tool selection annotation can be found below.
\input{body/prompts/tool_selection_annotation}

\paragraph{Tool Selection, Execution, and Response Processing.} 
These three steps are performed by the tool-using agent model based on the following prompts below. Basically, in both tool selection and execution, the tool descriptions generated by different methods are injected to the prompts in the corresponding section marked by \emph{tools\_info}.
After the response is processed, the result is fed into context of the next subtask as its context.

\input{body/prompts/tool_sel}

\input{body/prompts/param_gen}

\input{body/prompts/response_processing}

\paragraph{Evaluation Metrics}
For the tool-level F1 score, precision and recall are defined as
$\text{Prec} = \frac{TP}{TP + FP}, \quad
\text{Recall} = \frac{TP}{TP + FN}.$
A true positive (TP) corresponds to a tool that is both part of the ground truth and selected; a false positive (FP) is a selected tool that is not in the ground truth; and a false negative (FN) is a ground-truth tool that is not selected.

\subsection{StableToolBench Parameter Schema Correction}
\label{appendix:stb_preprocess}

This section describes the preprocessing procedure used to correct parameter schemas in StableToolBench.

We observe that a subset of tool parameter schemas in StableToolBench does not accurately reflect the true API requirements. Common issues include missing required parameters, inclusion of unsupported parameters, and incorrect parameter types. When such schemas are used during evaluation, tool invocations can fail with server-side errors, even when the model selects the correct tool and follows a reasonable call pattern. These failures introduce noise into benchmark results and confound comparisons between methods.

To address this issue, we connect StableToolBench to Smolagents~\cite{smolagents} and programmatically invoke each tool in an iterative manner. For each API, we examine server responses to identify mismatches between the declared schema and actual API behavior. Based on these observations, we revise parameter definitions to align with the true requirements enforced by the server, including parameter presence and type constraints.

By correcting these schema-level inconsistencies, we eliminate a class of evaluation failures that are unrelated to model capability. This preprocessing step improves the stability and interpretability of StableToolBench results and enables more reliable assessment of tool-use performance. Importantly, all methods --- including all baselines --- are evaluated on the corrected schemas, ensuring a fair comparison across methods. As a result, absolute numbers are not directly comparable to prior work that uses the original StableToolBench schemas. The prompt of the parameter fixing agent can be found below.
\input{body/prompts/preprocess}





\subsection{Training and Inference Details}
\label{subsec:train_and_infer}
All experiments are conducted on a single node equipped with 8 $\times$ NVIDIA A100 (80GB) GPUs.
For SFT, we develop based on the FSDP SFT Trainer of the verl library\footnote{\url{https://github.com/volcengine/verl}} to optimize training efficiency and streamline checkpoint saving and conversion.
The SFT hyperparameters are shown in Table~\ref{tab:hyperparams}.
For inference, we leverage vLLM~\citep{kwon2023efficient} for efficiency and perform top-p sampling with temperature 0.3, top-p 0.9, repetition\_penalty 1.1. 

We fine-tune Qwen3-4B-Instruct-2507\footnote{\url{https://huggingface.co/Qwen/Qwen3-4B-Instruct-2507}} due to its strong instruction following ability with $\mathcal{A}_{tr}$ of our synthesized dataset to make $\mathcal{A}_{ts}$ and the test queries unseen during training.

\input{tables/training_inference}

The prompts for Trace-Free+, Trace-Free at inference time and Trace-based Sample at training time for Trace-Free+ are shown below
\input{body/prompts/policy_trace_free}

\input{body/prompts/policy_trace_based}

\section{Case Study: Pattern Types Learned by Trace-Free+}
\label{appendix:case-study}

\subsection{Pattern Category Definitions}

Table~\ref{tab:pattern_category_def} defines the five interface pattern categories used throughout this analysis, with their level of application, key classification test, and a representative example.

\input{tables/pattern_category_def}

\subsection{Pattern Category Coverage}

This analysis is performed on tools from the Media and Finance categories of $\mathcal{A}_{ts}$, selected as representative domains with sufficient tool variety across all five pattern types. Observations from Table~\ref{tab:pattern_coverage}: Most notably, Trace-Free+ dramatically improves over D0 across all five categories, confirming that learned interface generation substantially enriches otherwise sparse baseline descriptions. While $D_0$ covers fewer than ~12\% of tools in any category, Trace-Free+ raises coverage to near-complete levels for tool selection scope (97.2\%) and parameter constraints (94.2\%), and introduces non-trivial gains even in harder categories such as cross-tool dependencies (30.0\%) and cross-parameter dependencies (17.1\%). This highlights that the model successfully internalizes and applies generalizable interface patterns from training, even without access to execution traces at inference time.

Compared to $D_1$ and $D_2$, Trace-Free+ shows a more nuanced trade-off. For tool selection scope and parameter constraints, all three methods achieve similarly high coverage, with $D_2$ slightly outperforming $D_1$ and Trace-Free+. For cross-tool dependencies, all methods converge to a similar range (30–32\%), suggesting this pattern is largely driven by explicit parameter guidance and is equally captured across approaches.

The main divergence appears in output description and cross-parameter dependencies. $D_1$ and $D_2$ achieve near-complete output description coverage (~99\%) because $D_1$'s data-independent documentation guidelines explicitly require enumerating output fields and their semantics directly from the schema --- making output description a near-guaranteed addition for every tool. \tracefreemix, by contrast, lags on this category (27.0\%), as it must infer output semantics from schema-level signals alone without the benefit of explicit guidelines, and tends to encode output information implicitly through usage scope rather than explicit field enumeration. This gap does not substantially affect downstream performance because output description primarily helps agents interpret tool responses, whereas the categories most critical for correct tool selection and argument construction --- tool selection scope and parameter constraints --- are the ones where \tracefreemix achieves its highest coverage (97.2\% and 94.2\% respectively). In contrast, Trace-Free+ outperforms both $D_1$ and $D_2$ on cross-parameter dependencies (17.1\% vs. 8–10\%), indicating a relative strength in modeling inter-parameter relationships from schema-level signals.

Overall, while $D_1$/$D_2$ achieve near-complete coverage on more surface-level documentation patterns, Trace-Free+ better captures certain structural constraints (e.g., parameter interactions), suggesting complementary strengths between template-based refinement and learned, schema-driven generalization.


\input{tables/pattern_category}

\subsection{Qualitative Examples}

To illustrate what the model learns to internalize, Table~\ref{tab:case-study} presents three examples of \tracefreemix-generated descriptions for unseen tools from $\mathcal{A}_{ts}$, highlighting the constraint types added without access to execution traces.

\input{tables/case_study}

\section{Additional Experimental Results}
\label{appendix:add_exp_results}



\noindent\textbf{Parameter Study: Ratio of Trace-free Data in Curriculum.}
Here, we also investigate the impact of the ratio of the trace-free data in the curriculum learning.
We fix the number of total training samples and investigate the impact of different learning curriculums.
%
%
To avoid overfitting, we limit the training to 2 epochs.
As we can observe from Table~\ref{tab:parameter_study}, the two-stage curriculum with 10\% trace-free data in the first stage and 90\% in the second stage is the most effective one. A hypothesis for this is that the second with 90\% trace-free is close to the trace-free scenario at inference time. We also tried 3 stages but the results are worse than 2 stages on StableToolBench.

\input{tables/trace_free_ratio}



\noindent\textbf{Transfer to BFCLv2.} Table~\ref{tab:bfcl-transfer} reports the full results of applying \tracefreemix-improved tool descriptions on BFCLv2~\citep{patil2025bfcl}.

\input{tables/bfcl_transfer}

\section{Future work}
This work opens several directions for future research. 
First, fine-tuning agents and improving tool interfaces can be done jointly for optimizing the performance of agents.
Second, within the scope of tool interface improvement, developing principled methods for query synthesis to cover both single-step and multi-step queries for both training and benchmarking is an interesting research direction.
%
Finally, while we focus on RESTful APIs, the same framework may apply to other domains such as databases or code execution environments. Finally, combining tool interface optimization with downstream agent training in an end-to-end fashion may lead to further gains.

%% file: body/prompts/annotator.tex
\begin{tcblisting}{title={Agentic Tool Annotator},colback=lightgrey,colframe=black,arc=1mm,boxrule=1pt,left=1mm,right=1mm,top=1mm,bottom=1mm,breakable,fontupper=\tiny\ttfamily,listing only,listing engine=listings,listing options={breaklines,breakautoindent=false,breakindent=0pt,keepspaces,tabsize=4,literate=
            {“}{"}{1}       
            {”}{"}{1}       
            {‘}{'}{1}       
            {’}{'}{1}       
            {—}{--}{1}      
            {–}{-}{1}       
            {±}{+/-}{2}     
            { }{ }{1}       
    }}
<system_prompt>
You are an agent that explores APIs to evaluate their health and discover working call patterns. You work in an Action and Observation loop until you return the final answer.

What you have
1) The JSON string of the current MCP schema for one API provider (initially).
2) A set of tools (APIs) defined by that schema that you can call to get concrete feedback.
3) Generic utilities that annotate the schema with health labels and successful call examples.

Your goal
- For each API in the provider, actively explore how to call it successfully.
- Use both the schema to infer true parameter names, types, and constraints.
- Adapt your calls when you see errors, instead of giving up after a single failure.
- After a reasonable amount of exploration for each API, decide its health and record at least one successful example call when possible.

What to annotate
- You must not change any tool name or delete tools.
- You should not rewrite the schema; instead, you infer how to call each API in practice.
- For each API, use `utility_annotate_health` to set:
  - `health = "good"` if you can find at least one meaningful, repeatable, successful call that returns plausible data.
  - `health = "bad"` if, after careful attempts and adapting to error messages, all calls fail due to issues you cannot fix from the client side (for example, persistent authorization errors, missing server configuration, 404 for the endpoint, or fundamentally broken behavior).
  - `health = "unknown"` if you cannot confidently determine whether the API works (for example, ambiguous or inconsistent errors, or not enough steps left to explore).
- When you mark an API as `"good"`, you should, whenever possible, also call `utility_annotate_example` to store one or more concrete successful call examples.
  - The `example` input must be a JSON string representing a list of argument objects, such as `[{"param1": "value1"}, {"param1": "value2", "param2": 3}]`.
  - Each object corresponds to a full set of arguments that you actually used in a successful call.
  - Prefer 1-3 diverse, minimal, safe examples that future agents can reuse directly.

How to explore APIs
- Prefer live Observations over assumptions.
- If a call fails with "Missing required parameter X" or similar, try again including that parameter.
- If a call fails with "Unexpected parameter Y" or type errors, adjust or remove that parameter.
- When behavior is unclear, run a minimal, low-risk test call to probe the interface rather than guessing wildly.
- Be efficient: you have a limited number of tool calls. Do not brute-force huge parameter spaces. Instead, reason about likely values based on descriptions.

Protocol
- Every step you take is an Action. An Action is a JSON blob of the form:
Action:
{
  "name": "<tool_name>",
  "arguments": <tool_input>
}
The "arguments" field must match what the tool expects. For most tools it is an object; for tools that accept a single value, it can be a raw value such as a string or number.
- Use the utility tools to write your findings back into the schema:
  - `utility_annotate_health`: Record the health label and a concise reason for each API.
  - `utility_annotate_example`: Record JSON examples of successful calls for APIs you understand well.
- After each Action, you will receive an Observation. Treat it as ground truth. The Observation may be plain text or structured JSON. Use it to decide the next Action.
- Use only listed tools. Do not invent tool names or parameters. Pass literal values, not variable names.

Required loop
1) Inspect the current schema to list all APIs you need to evaluate.
2) For each API, design and execute a small sequence of test calls to discover a working call (or to conclude that the API is broken or uncertain).
3) Update your calling strategy for that API whenever you observe new errors or unexpected behavior.
4) Once you have enough evidence, annotate the API's health and, if possible, record successful call examples using the utility tools.
5) Repeat until all APIs in the schema are annotated.
6) End by calling the "final_answer" tool with a concise summary of what you annotated.

Output and completion
- You must finish by calling the "final_answer" tool. It is the only way to complete the task.
- The "final_answer" tool will automatically validate and save the annotated schema.

Examples
Task: "Evaluate APIs and annotate their health"
Action:
{
  "name": "some_api",
  "arguments": {"parameter1": "value1"}
}
Observation: "{response: ...successful result...}"
Action:
{
  "name": "utility_annotate_health",
  "arguments": {"api_name": "some_api", "health": "good", "reason": "Successfully called with parameter1=value1 and obtained plausible results."}
}
Observation: "Health for API 'some_api' annotated."
Action:
{
  "name": "utility_annotate_example",
  "arguments": {"api_name": "some_api", "example": "[{\"parameter1\": \"value1\"}]"}
}
Observation: "Example for API 'some_api' annotated."
Action:
{
  "name": "final_answer",
  "arguments": "Annotated health and examples for all APIs."
}

Task: "Handle ambiguous or failing APIs"
Action:
{
  "name": "another_api",
  "arguments": {"paramA": "val"}
}
Observation: "Error: Missing required parameter 'paramB'"
Action:
{
  "name": "another_api",
  "arguments": {"paramA": "val", "paramB": 10}
}
Observation: "Error: 401 Unauthorized"
Action:
{
  "name": "utility_annotate_health",
  "arguments": {"api_name": "another_api", "health": "bad", "reason": "Repeated calls with corrected parameters still return server-side authorization errors."}
}
Observation: "Health for API 'another_api' annotated."
Action:
{
  "name": "final_answer",
  "arguments": "Annotated health as 'bad' for failing APIs where client-side fixes do not help."
}

Available tools
{%- for tool in tools.values() %}
- {{ tool.to_tool_calling_prompt() }}
{%- endfor %}

Rules you must follow
1) Always provide a tool call. If you are answering, call "final_answer".
2) Use only the arguments the tool expects. Pass literal values, not variable names.
3) Do not repeat an identical tool call with the exact same arguments.
4) Prefer evidence from Observations. If information is missing, probe with a minimal call.

Now Begin!
</system_prompt>

<user_prompt>
Evaluate and annotate the health of each API based on the schema by actively interacting with the tools.

The schema you are given is:
{{schema}}
</user_prompt>

\end{tcblisting}

%% file: tables/dataset_description.tex
\begin{table}[t]
\centering
\caption{Descriptive statistics of evaluation benchmarks. For StableToolBench, we only consider their solvable queries~\citep{guo2024stabletoolbench}.}
\label{tab:statistics}
\begin{tabular}{lccc}
\toprule
Dataset & Queries & Tools & Tools per Query\\
\midrule

\rowcolor{Gray}
\multicolumn{4}{c}{\textbf{RestBench}} \\
TMDB & 100 & 54 & 54\\
Spotify & 57 & 40 & 40\\

\midrule
\rowcolor{Gray}
\multicolumn{4}{c}{\textbf{StableToolBench}} \\

G1 Category & 153 & 364 & 4.21 \\
G1 Instruction & 163 & 820 & 5.29 \\
G1 Tool & 158 & 500 & 5.03 \\
G2 Category & 124 & 433 & 5.90\\
G2 Instruction & 105 & 595 & 6.49\\
G3 Instruction & 61 & 44 & 5.77 \\

\midrule
\rowcolor{Gray}
\multicolumn{4}{c}{\textbf{BFCLv2}} \\
Non-Live & 1390 & 1132 & 1.49\\
Live & 2251 & 739 & 2.96\\
\bottomrule
\end{tabular}
\end{table}

%% file: tables/synthetic_data_stats.tex
\begin{table}[t]
\centering
\caption{Statistics of the dataset we created in Sec~\ref{subsec:data-synthesis} for tool interface improvement.  To ensure evaluation on entirely unseen tools, all tools appearing as candidates in StableToolBench test queries are assigned to the test set $\mathcal{A}_{ts}$ and the remainder form $\mathcal{A}_{tr}$.}
\label{tab:synthetic_stats}
\begin{tabular}{lccc}
\toprule
Dataset & Queries & Tools & Tools per Query\\
\midrule

\rowcolor{Gray}
\multicolumn{4}{c}{\textbf{Synthesized SFT Data}} \\
$\mathcal{A}_{ts}$ & 4,726 & 4,585 & 6.81\\
$\mathcal{A}_{tr}$ & 2,189 & 991 & 9.21 \\
\bottomrule
\end{tabular}
\end{table}

%% file: body/prompts/tool_selection_annotation.tex
\begin{tcblisting}{title={Subtask Tool Selection Annotation},colback=lightgrey,colframe=black,arc=1mm,boxrule=1pt,left=1mm,right=1mm,top=1mm,bottom=1mm,breakable,fontupper=\tiny\ttfamily,listing only,listing engine=listings,listing options={breaklines,breakautoindent=false,breakindent=0pt,keepspaces,tabsize=4,literate=
            {“}{"}{1}       
            {”}{"}{1}       
            {‘}{'}{1}       
            {’}{'}{1}       
            {—}{--}{1}      
            {–}{-}{1}       
            {±}{+/-}{2}     
            { }{ }{1}       
    }}
<system_prompt>
You are an expert at analyzing task decomposition and determining whether subtasks require external API calls.
</system_prompt>

<user_prompt>
TASK: Analyze whether a subtask requires an API call or is just data processing.

CONTEXT:
- Original Query: {original_query}
- Subtask Input: {subtask_input}
- Previous Context: {previous_context}
- Available APIs: {tool_info}

INSTRUCTIONS:
1. Analyze the subtask input to understand what it's trying to accomplish
2. Consider whether the subtask needs to fetch NEW data from external sources
3. Determine if the subtask is just processing/analyzing data that's already available

CRITERIA for API NEED:
- The subtask needs to SEARCH for, FIND, GET, or RETRIEVE information
- The subtask needs to access external data sources
- The subtask cannot be completed with just the data from previous steps
- The subtask involves making requests to external services or databases

CRITERIA for NO API NEED (Processing Step):
- The subtask only processes/analyzes data from previous steps
- The subtask involves counting, filtering, selecting, comparing, or organizing existing data
- The subtask uses phrases like "from the list", "from the results", "based on the", "select one", etc.
- The subtask can be completed using only the information already gathered
- The subtask involves logical operations, calculations, or data manipulation on existing data

OUTPUT FORMAT:
Respond with a JSON object containing:
{{
    "needs_api": true/false,
    "reasoning": "Detailed explanation of your decision",
    "confidence": 0.0-1.0,
    "api_name": "Name of the API if needed, or empty string if not needed"
}}

EXAMPLES:
- Subtask: "Search for movies with Tom Hanks" -> needs_api: true, api_name: "search_movies"
- Subtask: "Count how many are comedies from the results" -> needs_api: false, api_name: ""
- Subtask: "Select the highest rated movie from the list" -> needs_api: false, api_name: ""
- Subtask: "Get movie details for the selected movie" -> needs_api: true, api_name: "get_movie_details"
- Subtask: "Compare the ratings of the top 3 movies" -> needs_api: false, api_name: ""
- Subtask: "Find similar movies to the selected one" -> needs_api: true, api_name: "get_similar_movies"

Now analyze the given subtask and provide your judgment. 
</user_prompt>

\end{tcblisting}

%% file: body/prompts/tool_sel.tex
\begin{tcblisting}{title={Subtask Tool Selection},colback=lightgrey,colframe=black,arc=1mm,boxrule=1pt,left=1mm,right=1mm,top=1mm,bottom=1mm,breakable,fontupper=\tiny\ttfamily,listing only,listing engine=listings,listing options={breaklines,breakautoindent=false,breakindent=0pt,keepspaces,tabsize=4,literate=
            {“}{"}{1}       
            {”}{"}{1}       
            {‘}{'}{1}       
            {’}{'}{1}       
            {—}{--}{1}      
            {–}{-}{1}       
            {±}{+/-}{2}     
            { }{ }{1}       
    }}
<system_prompt>
You are an expert API selector. Given a user query for a specific subtask and available tools/APIs, you need to select exactly ONE most appropriate API to handle this subtask.
</system_prompt>

<user_prompt>
Available Tools and APIs:
{tools_info}

Your task:
1. Analyze the subtask query and the "subtask_output" in the Context Section to understand what specific information is needed
2. Select exactly ONE API from the available tools that is most appropriate for this subtask
3. Focus on the current subtask only - don't consider future steps

### Context Section
{context_section}

### Subtask Query
{query}

Important: You must select exactly ONE API that is most appropriate for this specific subtask.

You must respond in JSON format with exactly one selected API:
{{
    "selected_api": {{
        "reasoning": "why this specific API was selected for this subtask",
        "api_name": "api_name_here",
    }}
}} 
</user_prompt>

\end{tcblisting}

%% file: body/prompts/param_gen.tex
\begin{tcblisting}{title={Subtask Tool Execution (Parameter Generation)},colback=lightgrey,colframe=black,arc=1mm,boxrule=1pt,left=1mm,right=1mm,top=1mm,bottom=1mm,breakable,fontupper=\tiny\ttfamily,listing only,listing engine=listings,listing options={breaklines,breakautoindent=false,breakindent=0pt,keepspaces,tabsize=4,literate=
            {“}{"}{1}       
            {”}{"}{1}       
            {‘}{'}{1}       
            {’}{'}{1}       
            {—}{--}{1}      
            {–}{-}{1}       
            {±}{+/-}{2}     
            { }{ }{1}       
    }}
<system_prompt>
You are a helpful assistant that generates parameters for an API call.
</system_prompt>

<user_prompt>
Given a subtask and an API and its description, you need to first write your reasoning step by step in plain text about how to extract the correct parameters. After reasoning, you must then output the final parameters in strict JSON format according to the API description.

Please note that:

The API description can help you better understand the use of the API.

Ensure the parameters you output are correct. The output must contain the required parameters, and may contain the optional parameters if needed. If no parameters exist in the required and optional parameters, just leave it as {{"Parameters":{{}}}}.

If the subtask mentions other APIs, you should ONLY consider the API description I give and do not consider other APIs.

Parameter Extraction from Previous Context: When the API requires path parameters (like person_id, movie_id, tv_id, company_id, etc.), you may have to extract them from the subtask_output of previous steps if they are missing from the subtask input. Try to extract the numeric ID values from these text descriptions and use them as the corresponding path parameters.

You must ONLY output in a parsable JSON format for the final answer, with no extra explanations, notes, or comments after it.

The output must have two parts:

"Reasoning": your step-by-step reasoning as plain text.

"Parameters": the final extracted parameters in JSON format.

An example output looks like:

{{
  "Reasoning": "The subtask asks for person details. The required parameter is person_id. From previous_log, I see that person_id is 190. Therefore, the correct parameter is person_id=190.",
  "Parameters": {{
    "person_id": 190
  }}
}}

There are logs of previous questions and answers:
{previous_log}

This is API tool documentation:
{api_instruction}

This is the current subtask:
{question}

Output:
</user_prompt>

\end{tcblisting}

%% file: body/prompts/response_processing.tex
\begin{tcblisting}{title={Subtask Tool Response Processing},colback=lightgrey,colframe=black,arc=1mm,boxrule=1pt,left=1mm,right=1mm,top=1mm,bottom=1mm,breakable,fontupper=\tiny\ttfamily,listing only,listing engine=listings,listing options={breaklines,breakautoindent=false,breakindent=0pt,keepspaces,tabsize=4,literate=
            {“}{"}{1}       
            {”}{"}{1}       
            {‘}{'}{1}       
            {’}{'}{1}       
            {—}{--}{1}      
            {–}{-}{1}       
            {±}{+/-}{2}     
            { }{ }{1}       
    }}
<system_prompt>
You are a helpful assistant.
</system_prompt>

<user_prompt>
You should answer the question based on the response output by the API tool.

Please note that:
1. Try to organize the response into a natural language answer.
2. We will not show the API response to the user, thus you need to make full use of the response and give the information in the response that can satisfy the user's question in as much detail as possible.
3. The question may have dependencies on answers of other questions, so we will provide logs of previous questions and answers.

There are logs of previous questions and answers: 
{context_section}

This is the user's question: {subtask query}

This is the response output by the API tool: 
{call_result}
...
</user_prompt>

\end{tcblisting}

%% file: body/prompts/preprocess.tex
\begin{tcblisting}{title={Schema Parameter Fixing},colback=lightgrey,colframe=black,arc=1mm,boxrule=1pt,left=1mm,right=1mm,top=1mm,bottom=1mm,breakable,fontupper=\tiny\ttfamily,listing only,listing engine=listings,listing options={breaklines,breakautoindent=false,breakindent=0pt,keepspaces,tabsize=4,literate=
            {“}{"}{1}       
            {”}{"}{1}       
            {‘}{'}{1}       
            {’}{'}{1}       
            {—}{--}{1}      
            {–}{-}{1}       
            {±}{+/-}{2}     
            { }{ }{1}       
    }}
<system_prompt>
You are an agent that rewrites a tool's schema so future tool calls succeed. You work in an Action and Observation loop until you return the final answer.

What you have
1) The JSON string of the current schema that you will rewrite (initially).
2) A set of tools defined by that schema that you can call to get concrete feedback.
3) Generic utilities that help edit the schema incrementally.
4) The interactive history of past tool calls and their results.

What to change
- You may change a tool's description and parameters.
- You must not change any tool name. You must not change the structure of the schema.
- Add types, defaults, value ranges, enums, and constraints when the history shows they are needed.
- Make hidden requirements explicit in the description and parameter docs.
- Rewrite the API provider description and API descriptions to be clear and helpful for future LLM function calls.

How to write the description
Start with a one-sentence plain summary of what the tool does and the problem it solves. Then document:
- Inputs: each parameter with type, required/optional, default, min/max, allowed values, and formatting rules.
- Data model: shape of nested objects or arrays; limits on size or length; paging or cursors if present.
- Outputs: what the tool returns and what it does not return, including common items a caller might expect but are not provided.
- Primary use cases: the common ways the tool is used.
- Non-use cases: when the tool should not be used.

How to decide changes
- Prefer observed behavior from the history over assumptions. If a call failed with "Missing required parameter X," make X required and document it.
- If a call failed with "Unexpected parameter Y," remove or rename Y to match the actual interface.
- When behavior is unclear, run a minimal tool call to probe the interface rather than guessing.
- Keep backward compatibility with prior successful calls when possible.

Protocol
- Every step you take is an Action. An Action is a JSON blob of the form:
  Action:
  {
    "name": "<tool_name>",
    "arguments": <tool_input>
  }
  The "arguments" field must match what the tool expects. For most tools it is an object; for tools that accept a single value, it can be a raw value such as a string or number.
- Use the utility tools to update the schema incrementally.
  - `utility_update_provider_description`: Update the provider's description.
  - `utility_update_api_description`: Update an API's description.
  - `utility_update_api_parameters`: Update an API's parameters (JSON string).
  - `utility_remove_api_parameters`: Remove the 'parameters' field from an API (making it accept no parameters).
  - `utility_print_schema`: See the current state of the schema.
- After each Action, you will receive an Observation. Treat it as ground truth. The Observation may be plain text or structured JSON. Use it to decide the next Action.
- Use only listed tools. Do not invent tool names or parameters. Pass literal values, not variable names.

Required loop
1) Inspect the current schema and the history.
2) Exercise the tool(s) with targeted calls to reveal true parameter names, required fields, and constraints.
3) Edit the schema incrementally using the utility tools to reflect observed behavior.
4) Validate by re-running the previously failing calls until they succeed or until you reach the real limits of the tool.
5) End with the final answer tool.

Output and completion
- You must finish by calling the "final_answer" tool. It is the only way to complete the task.
- The "final_answer" tool will automatically save the schema.

Examples
Task: "Rewrite the schema based on the log"
Action:
{
  "name": "some_tool_in_schema",
  "arguments": {"parameter1": "value1"}
}
Observation: "TypeError: some_tool_in_schema() takes 0 positional arguments but 1 was given"
Action:
{
  "name": "utility_remove_api_parameters",
  "arguments": {"api_name": "some_tool_in_schema"}
}
Observation: "Parameters field removed from API 'some_tool_in_schema'."
Action:
{
  "name": "some_tool_in_schema",
  "arguments": {}
}
Observation: "{response: ...}"
Action:
{
  "name": "final_answer",
  "arguments": "The schema has been rewritten."
}

Task: "Update parameters"
Action:
{
  "name": "another_tool",
  "arguments": {"paramA": "val"}
}
Observation: "Missing required parameter: paramB"
Action:
{
  "name": "utility_update_api_parameters",
  "arguments": {
    "api_name": "another_tool", 
    "parameters_json": "{\"paramA\": {\"type\": \"string\", \"required\": true, \"description\": \"...\"}, \"paramB\": {\"type\": \"string\", \"required\": true, \"description\": \"...\"}}"
  }
}
Observation: "Parameters for API 'another_tool' updated."
Action:
{
  "name": "final_answer",
  "arguments": "Updated parameters."
}

Task: "Clarify API description"
Action:
{
  "name": "utility_update_api_description",
  "arguments": {
    "api_name": "complex_tool",
    "description": "This tool calculates the mortgage payment. Inputs: 'principal' (number, required), 'rate' (number, required, annual interest rate in percentage), 'years' (number, required, loan term). Output: Monthly payment amount."
  }
}
Observation: "Description for API 'complex_tool' updated."
Action:
{
  "name": "final_answer",
  "arguments": "Updated API description to be more clear."
}

Available tools
{%- for tool in tools.values() %}
- {{ tool.to_tool_calling_prompt() }}
{%- endfor %}

Rules you must follow
1) Always provide a tool call. If you are answering, call "final_answer".
2) Use only the arguments the tool expects. Pass literal values, not variable names.
3) Do not repeat an identical tool call with the exact same arguments.
4) Prefer evidence from Observations and the history over guesses. If information is missing, probe with a minimal call.

Now Begin!
</system_prompt>

<user_prompt>
Rewrite the schema of the tool based on the log below and interacting with the tools.

The schema you are given is:
{{schema}}
</user_prompt>

\end{tcblisting}

%% file: tables/training_inference.tex
\begin{table}[htbp]
\centering
\small
\begin{tabular}{lc}
\toprule[1.5pt]
\textbf{Parameter} & \textbf{SFT} \\ \midrule
Base Model              & Qwen3-4B-Instruct-2507 \\
Hardware                & 1 $\times$ 8-A100 GPU Node \\
Optimizer               & AdamW \\
Learning Rate           & $5.0 \times 10^{-5}$ \\
LR Scheduler            & Cosine \\
Training Epochs         & 2.0 \\
LoRA Rank               & 64 \\
Precision               & bf16 \\
Max Length              & 2{,}048 \\
Effective Batch Size    & 8 \\
DeepSpeed Stage         & ZeRO-3 \\
\bottomrule[1.5pt]
\end{tabular}
\caption{Hyperparameters for SFT.}
\label{tab:hyperparams}
\end{table}

%% file: body/prompts/policy_trace_free.tex
\begin{tcblisting}{title={Prompt for Trace-Free+ (Inference and Trace-free Samples in Training) and Trace-Free},colback=lightgrey,colframe=black,arc=1mm,boxrule=1pt,left=1mm,right=1mm,top=1mm,bottom=1mm,breakable,fontupper=\tiny\ttfamily,listing only,listing engine=listings,listing options={breaklines,breakautoindent=false,breakindent=0pt,keepspaces,tabsize=4,literate=
            {“}{"}{1}       
            {”}{"}{1}       
            {‘}{'}{1}       
            {’}{'}{1}       
            {—}{--}{1}      
            {–}{-}{1}       
            {±}{+/-}{2}     
            { }{ }{1}       
    }}
<system_prompt>
You are an API documentation specialist.
</system_prompt>

<user_prompt>

Rewrite the API description so an AI agent can:
1) Decide when to use this API
2) Generate valid parameters

Inputs:
- API name: {tool_name}
- Parameter schema: {parameter_json}
- Baseline description: {original_description}

Infer (do not output):
- When to use vs not use this API
- Required vs optional parameters
- Parameter meanings and constraints
- Cross-parameter dependencies or exclusions
- Common parameter mistakes
  - no examples are provided, infer from the schema and baseline description only

Write a clear API description that:
- States when to use and NOT use the API
- Does not invent or reference non-provided APIs
- Explains each parameter's meaning, type, required/optional status, constraints, and defaults
- Describes likely validation failures and how to avoid them
- Abstracts patterns into general rules
- Does not restate the full schema verbatim
- Does not mention whether examples were provided

You may replace the baseline description entirely.

Output ONLY valid JSON (no markdown, no code blocks):
{{"description": "<your improved API description here>"}}

</user_prompt>

\end{tcblisting}

%% file: body/prompts/policy_trace_based.tex
\begin{tcblisting}{title={Prompt for Trace-Free+ (Trace-based Samples in Training)},colback=lightgrey,colframe=black,arc=1mm,boxrule=1pt,left=1mm,right=1mm,top=1mm,bottom=1mm,breakable,fontupper=\tiny\ttfamily,listing only,listing engine=listings,listing options={breaklines,breakautoindent=false,breakindent=0pt,keepspaces,tabsize=4,literate=
            {“}{"}{1}       
            {”}{"}{1}       
            {‘}{'}{1}       
            {’}{'}{1}       
            {—}{--}{1}      
            {–}{-}{1}       
            {±}{+/-}{2}     
            { }{ }{1}       
    }}
<system_prompt>
You are an API documentation specialist.
</system_prompt>

<user_prompt>
Rewrite the API description so an AI agent can:
1) Decide when to use this API
2) Generate valid parameters

Inputs:
- API name: {tool_name}
- Parameter schema: {parameter_json}
- Example queries + errors: {query_examples}
- Baseline description: {original_description}

Infer (do not output):
- When to use vs not use this API
- Common parameter mistakes
- Required vs optional parameters
- Cross-parameter constraints

Write a clear API description that:
- States when to use and NOT use the API
- Does not invent other APIs
- Explains each parameter's meaning, type, required/optional status, constraints, and defaults
- Describes common validation failures and how to avoid them
- Abstracts examples into general rules
- Does not restate the full schema or copy examples

You may replace the baseline entirely.

Output ONLY valid JSON (no markdown, no code blocks):
{{"description": "<your improved API description here>"}}

</user_prompt>

\end{tcblisting}

%% file: tables/pattern_category_def.tex
\begin{table*}[h]
\centering
\small
\begin{tabular}{clp{2.5cm}p{2.5cm}p{2.5cm}}
\toprule
\textbf{\#} & \textbf{Category} & \textbf{Level / Key Test} & \textbf{Explanation} & \textbf{Example} \\
\midrule
1 & Tool selection scope & Tool-level. Does it help an agent decide between multiple options? &
Explicitly states when to use this tool vs.\ another, when NOT to use it, or compares it to a similar tool. Must go beyond a simple purpose statement. &
``Use this API when you need to retrieve a list of artworks by search query. Do not use it if you require detailed metadata such as dimensions or provenance. For more detail, use Detect Features instead.'' \\
\addlinespace
2 & Cross-tool dependencies & Tool-level. Does it name a specific upstream endpoint? &
A parameter value must come from calling a specific named tool or endpoint first. The description explicitly names the other tool/endpoint --- not just general prior state. &
``The \texttt{hash} parameter must be extracted from the Check Status API response when \texttt{movie\_status} equals DONE, and \texttt{vsid} must come directly from the Start Movie Session API response.'' \\
\addlinespace
3 & Output description & Tool-level. Does it enumerate response fields or explicitly state what is NOT returned? &
Explicitly enumerates response fields, data types, or structure --- or explicitly states what is not included. Does not apply to general purpose statements that merely describe what the tool retrieves. &
``Returns all static metadata including logo, description, official website URL, and social links. Does not include videos, subcategories, or category-specific details.'' \\
\addlinespace
4 & Parameter constraints & Parameter-level. Does it restrict valid values or structure? &
Restricts what values a parameter can take or specifies how it must be structured: enums, numeric ranges, case sensitivity rules, date formats, encoding schemes, separator conventions, or array structure. &
``Must be one of \texttt{en} (English) or \texttt{nl} (Dutch). \texttt{limit} must be a positive integer between 1 and 100. \texttt{date} must be in \texttt{YYYY-MM-DD} format. \texttt{image} must be a Base64-encoded string.'' \\
\addlinespace
5 & Cross-param dependencies & Parameter-level. Does it constrain one parameter based on another? &
Constraints between parameters within the same tool --- parameters that must be paired together or are mutually exclusive. &
``\texttt{longitude} (requires \texttt{lat}).'' \\
\bottomrule
\end{tabular}
\caption{Definitions of the five interface pattern categories used to classify tool description improvements. Categories 1--3 are tool-level (apply to the description as a whole); Categories 4--5 are parameter-level (apply per parameter).}
\label{tab:pattern_category_def}
\end{table*}

%% file: tables/pattern_category.tex
\begin{table}[h]
  \centering
  \begin{tabular}{clrrrr}
  \toprule
  \textbf{\#} & \textbf{Category} & \textbf{$D_0$} & \textbf{$D_1$} & \textbf{$D_2$} & \textbf{Trace-Free+} \\
  \midrule
  1 & Tool selection scope     & 3.3\%  & 99.7\% & 100.0\% & 97.2\% \\
  2 & Cross-tool dependencies  & 0.3\%  & 32.0\% &  21.1\% & 30.0\% \\
  3 & Output description       & 11.6\% & 98.6\% & 100.0\% & 27.0\% \\
  4 & Parameter constraints    & 9.9\%  & 87.9\% &  89.5\% & 94.2\% \\
  5 & Cross-param dependencies & 0.3\%  &  8.5\% &  21.1\% & 17.1\% \\
  \bottomrule
  \end{tabular}
  \caption{Percentage of tool descriptions covering each of the five interface pattern categories, comparing $D_0$, $D_1$, $D_2$, and \tracefreemix. $D_0$ descriptions are consistently sparse across all categories; all three improved variants close most of the gap on scope and parameter constraints, with differences concentrated on cross-tool dependencies and output description.}
  \label{tab:pattern_coverage}
\end{table}

%% file: tables/case_study.tex
\begin{table*}[t]
  \centering
  \small
  \begin{tabular}{p{0.11\linewidth} p{0.38\linewidth} p{0.38\linewidth}}
  \toprule
   & \textbf{$D_0$ (Original)} & \textbf{\tracefreemix (Ours)} \\
  \midrule

  \textbf{API} & \multicolumn{2}{l}{\texttt{Walk Score} (provider: \texttt{Walk Score})} \\
  \addlinespace
  \textbf{Parameters} & \multicolumn{2}{l}{\texttt{lat} (str), \texttt{lon} (str), \texttt{address} (str), \texttt{wsapikey} (str), \texttt{bike} (str, opt.), \texttt{transit} (str, opt.), \texttt{format} (str, opt.)} \\
  \midrule
  \textbf{Description} &
  Get Walk Score. \texttt{address}: ``The URL encoded address.''
  \texttt{lat}/\texttt{lon}: no range stated.
  \texttt{format}: ``Type of result to return: (movie, series, episode).'' &
  Retrieves Walk, Transit, and Bike scores for a location.
  \texttt{lat} must be a decimal in $[-90, 90]$; \texttt{lon} in $[-180, 180]$.
  \texttt{address} must \emph{not} be URL-encoded.
  \texttt{bike} and \texttt{transit} must be exactly \texttt{'1'} --- not \texttt{'true'} or \texttt{'on'}.
  \texttt{format} must be \texttt{'json'} or empty (XML default); any other value is rejected.
  Do not use for real-time traffic data or historical trends. \\
  \addlinespace
  \textbf{Patterns} &
  --- &
  \textit{Parameter constraints} (\texttt{lat}/\texttt{lon} numeric ranges; exact-value flags for \texttt{bike}, \texttt{transit}, \texttt{format}; format correction for \texttt{address});
  \textit{Tool selection scope} (exclusions: not for real-time traffic or historical trends) \\

  \midrule\midrule

  \textbf{API} & \multicolumn{2}{l}{\texttt{Player Trophy Information} (provider: \texttt{Trackmania})} \\
  \addlinespace
  \textbf{Parameters} & \multicolumn{2}{l}{\texttt{player\_id} (str), \texttt{limit} (float, opt.), \texttt{page} (float, opt.)} \\
  \midrule
  \textbf{Description} &
  Get the trophy info of a player.
  \texttt{limit}: ``The result limit that gets returned. Default limit is \texttt{50}.''
  \texttt{page}: ``The result page, each containing 50 results. Default page is \texttt{0}.'' &
  Retrieves a player's trophy records, paginated.
  \texttt{player\_id} must be obtained from the \texttt{/players} endpoint; invalid or missing IDs cause validation failure.
  \texttt{limit} must be between \textbf{1 and 100 inclusive}; values above 100 or below 1 are rejected.
  \texttt{page} must be a \textbf{non-negative integer}; negative or non-integer inputs are invalid.
  Do not use for general player statistics or game metadata. \\
  \addlinespace
  \textbf{Patterns} &
  --- &
  \textit{Parameter constraints} (\texttt{limit} $\in [1,100]$; \texttt{page} $\geq 0$);
  \textit{Cross-tool dependencies} (\texttt{player\_id} must come from \texttt{/players} endpoint);
  \textit{Tool selection scope} (trophy records only, not general player stats) \\

  \midrule\midrule

  \textbf{API} & \multicolumn{2}{l}{\texttt{GET queue} (provider: \texttt{League of Legends API})} \\
  \addlinespace
  \textbf{Parameters} & \multicolumn{2}{l}{\texttt{region} (str), \texttt{queue} (str), \texttt{division} (str), \texttt{tier} (str)} \\
  \midrule
  \textbf{Description} &
  Returns the list of people in division, tier and according to the type of queue.
  All four parameter descriptions are empty. &
  Retrieves ranked-queue player listings by division, tier, and region.
  \texttt{division} must be one of \texttt{'I'}, \texttt{'II'}, \texttt{'III'}, \texttt{'IV'}.
  \texttt{queue} must be a valid ranked queue identifier (e.g., \texttt{'RANKED\_SOLO\_5x5'}).
  \texttt{region} must be a supported server code (e.g., \texttt{'NA'}, \texttt{'EUW'}, \texttt{'KR'}).
  \texttt{tier} must be a valid rank tier (e.g., \texttt{'Gold'}, \texttt{'Diamond'}).
  All four parameters are required; any missing or malformed value causes a validation error. \\
  \addlinespace
  \textbf{Patterns} &
  --- &
  \textit{Parameter constraints} (enumerated valid values for all four parameters, filled from empty stubs);
  \textit{Tool selection scope} (ranked queue listings only, not live match data) \\

  \bottomrule
  \end{tabular}
  \caption{Three examples of \tracefreemix-generated descriptions for unseen tools compared against $D_0$. In all cases the model infers constraints without execution traces.}
  \label{tab:case-study}
\end{table*}

%% file: tables/trace_free_ratio.tex
\begin{table*}[]
\caption{Parameter Study for Trace-Free+ in Trace-free Evaluation}
\label{tab:parameter_study}
\resizebox{\textwidth}{!}{%
\begin{tabular}{lcccccccccccc}
\toprule
& \multicolumn{12}{c}{\textbf{StableToolBench}} \\
\cmidrule(lr){2-13}

& \multicolumn{2}{c}{G1 Category}
& \multicolumn{2}{c}{G1 Instruction}
& \multicolumn{2}{c}{G1 Tool}
& \multicolumn{2}{c}{G2 Category}
& \multicolumn{2}{c}{G2 Instruction}
& \multicolumn{2}{c}{G3 Instruction} \\
\cmidrule(lr){2-3} \cmidrule(lr){4-5} \cmidrule(lr){6-7}
\cmidrule(lr){8-9} \cmidrule(lr){10-11} \cmidrule(lr){12-13}
 Ratios
& SL & QL
& SL & QL
& SL & QL
& SL & QL
& SL & QL
& SL & QL 
\\


\midrule
(0.1, 0.9) & 
73.8 ± 0.7 & 65.6 ± 1.0 & 72.6 ± 1.5 & 60.0 ± 1.0 & 70.0 ± 1.4 & 56.4 ± 1.6 & 68.7 ± 0.7 & 47.5 ± 0.8 & 71.4 ± 1.1 & 48.3 ± 1.6 & 63.9 ± 1.8 & 46.4 ± 4.1 
\\
(0.3, 0.7) &	 
73.7 ± 1.9 & 63.6 ± 2.8 & 73.2 ± 2.3 & 61.3 ± 4.3 & 68.3 ± 1.1 & 52.8 ± 0.7 & 68.6 ± 0.0 & 47.9 ± 0.6 & 69.1 ± 2.9 & 45.4 ± 3.6 & 61.1 ± 3.0 & 41.8 ± 3.5
\\
(0.5, 0.5) & 75.8 ± 0.0 & 66.0 ± 0.0 & 71.1 ± 0.0 & 64.3 ± 0.0 & 69.2 ± 0.0 & 55.6 ± 0.0 & 65.9 ± 0.0 & 48.3 ± 0.0 & 68.9 ± 0.0 & 44.9 ± 0.0 & 59.8 ± 0.0 & 39.3 ± 0.0 \\



\\
\tracefree &  
71.8 ± 2.3 & 62.7 ± 2.1 & 70.8 ± 0.1 & 60.8 ± 0.9 & 68.5 ± 2.7 & 53.6 ± 2.8 & 66.4 ± 1.2 & 44.1 ± 0.0 & 69.2 ± 4.0 & 46.4 ± 3.6 & 59.9 ± 1.2 & 41.8 ± 1.2 \\

\bottomrule
\end{tabular}%
}
\end{table*}

%% file: tables/bfcl_transfer.tex
\begin{table}[t]
\centering
\caption{Transfer evaluation on BFCLv2. We apply \tracefreemix to improve tool descriptions and evaluate on the Non-Live and Live splits.}
\label{tab:bfcl-transfer}
\scriptsize
\setlength{\tabcolsep}{6pt}
\begin{tabular}{lcccc}
\toprule
& \multicolumn{2}{c}{\textbf{Non-Live}}
& \multicolumn{2}{c}{\textbf{Live}} \\
\cmidrule(lr){2-3} \cmidrule(lr){4-5}

\textbf{Model}
& D0 & + \tracefreemix
& D0 & + \tracefreemix \\
\midrule

GPT-4.1
& 88.56 & \textbf{89.04}
& 79.28 & \textbf{80.63} \\

Claude Sonnet 4.5
& 63.11 & \textbf{65.50}
& 52.40 & \textbf{53.45} \\

Gemini-3-pro-preview
& 91.24 & \textbf{91.55}
& 84.98 & \textbf{86.41} \\

\bottomrule
\end{tabular}
\end{table}